\documentclass[conference]{IEEEtran}
\IEEEoverridecommandlockouts
\usepackage{cite}
\usepackage{amsmath,amssymb,amsfonts}
\usepackage{algorithmic}
\usepackage{graphicx}
\usepackage{textcomp}
\usepackage{xcolor}
\usepackage{booktabs}
\usepackage{multirow}
\usepackage{url}
\usepackage{hyperref}
\usepackage{placeins}
\usepackage{float}
\usepackage{capt-of}
\usepackage{dsfont}
\def\BibTeX{{\rm B\kern-.05em{\sc i\kern-.025em b}\kern-.08em
    T\kern-.1667em\lower.7ex\hbox{E}\kern-.125emX}}
\begin{document}

% \title{Interpretable Alzheimer’s Diagnosis via Multimodal Fusion of Regional Brain Experts\\
% {\footnotesize \textsuperscript{*}Note: Sub-titles are not captured for https://ieeexplore.ieee.org  and
% should not be used}

\title{Interpretable Alzheimer’s Disease Diagnosis via \\ Multimodal Fusion of Regional Brain Experts
%\thanks{Identify applicable funding agency here. If none, delete this.}
}

% \author{\IEEEauthorblockN{1\textsuperscript{st} Given Name Surname}
% \IEEEauthorblockA{\textit{dept. name of organization (of Aff.)} \\
% \textit{name of organization (of Aff.)}\\
% City, Country \\
% email address or ORCID}
% \and
% \IEEEauthorblockN{2\textsuperscript{nd} Given Name Surname}
% \IEEEauthorblockA{\textit{dept. name of organization (of Aff.)} \\
% \textit{name of organization (of Aff.)}\\
% City, Country \\
% email address or ORCID}
% \and
% \IEEEauthorblockN{3\textsuperscript{rd} Given Name Surname}
% \IEEEauthorblockA{\textit{dept. name of organization (of Aff.)} \\
% \textit{name of organization (of Aff.)}\\
% City, Country \\
% email address or ORCID}
% \and
% \IEEEauthorblockN{4\textsuperscript{th} Given Name Surname}
% \IEEEauthorblockA{\textit{dept. name of organization (of Aff.)} \\
% \textit{name of organization (of Aff.)}\\
% City, Country \\
% email address or ORCID}
% \and
% \IEEEauthorblockN{5\textsuperscript{th} Given Name Surname}
% \IEEEauthorblockA{\textit{dept. name of organization (of Aff.)} \\
% \textit{name of organization (of Aff.)}\\
% City, Country \\
% email address or ORCID}
% \and
% \IEEEauthorblockN{6\textsuperscript{th} Given Name Surname}
% \IEEEauthorblockA{\textit{dept. name of organization (of Aff.)} \\
% \textit{name of organization (of Aff.)}\\
% City, Country \\
% email address or ORCID}
% }

\author{
\IEEEauthorblockN{Farica Zhuang\textsuperscript{*}, Shu Yang\textsuperscript{*}}
\IEEEauthorblockA{\textit{University of Pennsylvania}\\
Philadelphia, USA \\
\{farica, syang11\}@upenn.edu}
\and
\IEEEauthorblockN{Dinara Aliyeva}
\IEEEauthorblockA{\textit{University of North Carolina}\\
Chapel Hill, USA \\
adinara@cs.unc.edu}
\and
\IEEEauthorblockN{Zixuan Wen, Duy Duong-Tran, Christos Davatzikos}
\IEEEauthorblockA{\textit{University of Pennsylvania}\\
Philadelphia, USA \\
\{zxwen, duongtrd, chdavatz\}@upenn.edu}
\and 
% \IEEEauthorblockN{Christos Davatzikos}
% \IEEEauthorblockA{\textit{University of Pennsylvania}\\
% Philadelphia, USA \\
% chdavatz@upenn.edu}
% \and
\IEEEauthorblockN{Tianlong Chen}
\IEEEauthorblockA{\textit{University of North Carolina}\\
Chapel Hill, USA \\
tianlong@cs.unc.edu}
\and
\IEEEauthorblockN{Song Wang}
\IEEEauthorblockA{\textit{University of Central Florida}\\
Orlando, USA \\
song.wang@ucf.edu}
\and
\IEEEauthorblockN{Li Shen\textsuperscript{**}}
\IEEEauthorblockA{\textit{University of Pennsylvania}\\
Philadelphia, USA \\
lishen@upenn.edu}

\thanks{\textsuperscript{*}These authors contributed equally to this work.}
\thanks{\textsuperscript{**}Correspondence to: lishen@upenn.edu}
}

\maketitle

\vspace{-20pt}

\begin{abstract}
Accurate and early diagnosis of Alzheimer's disease (AD) is critical for effective intervention and requires integrating complementary information from multimodal neuroimaging data.
% can benefit from integrating complementary information from multiple modalities, mirroring clinical practice. 
However, conventional fusion approaches often rely on simple concatenation of features, which cannot adaptively balance the contributions of biomarkers such as amyloid PET and MRI across brain regions. In this work, we propose MREF-AD, a Multimodal Regional Expert Fusion model for AD diagnosis. It is a Mixture-of-Experts (MoE) framework that models mesoscopic brain regions within each modality as independent experts and employs a gating network to learn subject-specific fusion weights. Utilizing tabular neuroimaging and demographic information from the Alzheimer's Disease Neuroimaging Initiative (ADNI), MREF-AD achieves competitive performance over strong classic and deep baselines 
while providing interpretable, modality- and region-level insight into how structural and molecular imaging jointly contribute to AD diagnosis. The source code is available at \url{https://github.com/PennShenLab/mref-ad}.
\end{abstract}


\section{Introduction}
Alzheimer's disease is a progressive neurodegenerative disorder, leading to cognitive decline and dementia \cite{selkoe2001alzheimer, jack2010hypothetical}. Despite the importance of early and accurate diagnosis for effective intervention, it remains challenging because clinical symptoms may overlap with healthy aging and other dementias. As such, in clinical practice, diagnosis often requires integrating multiple sources of evidence \cite{jack2018nia, frisoni2010clinical}. This motivates the development of computational models that can combine complementary modalities for more robust and early AD detection. 

Two commonly used imaging modalities in AD diagnosis are amyloid positron emission tomography (PET) and magnetic resonance imaging (MRI), which provide molecular and structural information of the brain, respectively. These modalities offer complementary information of two distinct yet related biological processes. In particular, amyloid PET captures the burden of amyloid-$\beta$ plaques in the brain, while MRI captures brain volume and is used to measure structural atrophy across cortical and subcortical regions. Both amyloid-$\beta$ plaques and brain volume are established markers that are widely associated with AD \cite{jagust2018imaging, zhang2011multimodal, frisoni2010clinical, jack2010hypothetical}. In clinical and large-scale studies such as ADNI~\cite{petersen2010alzheimer}, neuroimaging scans are commonly also summarized as their quantitative measurements in tabular format for specific regions of interests (ROIs), providing fine-grained, high-dimensional anatomical details. Measurements are quantified in standardized uptake value ratios (SUVR) of amyloid-$\beta$ plaques for PET and regional brain volumes for MRI \cite{jagust2024adni-pet-core-at-20, jack2024overview-of-adni-mri}. 

\begin{figure*}[tb!]
\centerline{
  \includegraphics[width=0.8\textwidth]
  {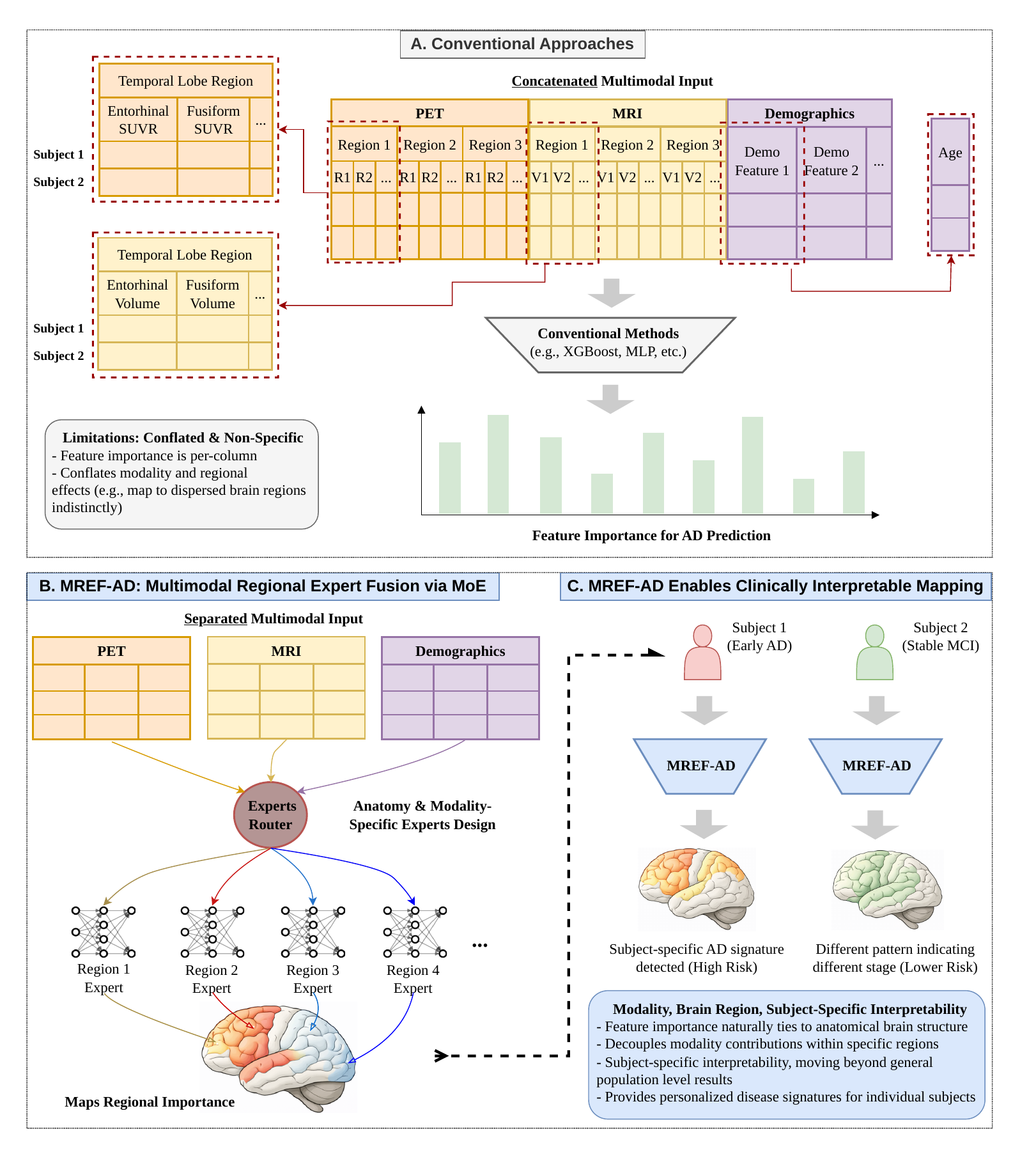} %Figures/LLM_bias_fig-fig1_v2-cropped.pdf} 
}
  \caption{A comparison of strategies for multimodal Alzheimer's study. \textbf{A.} Conventional methods usually adopt early-fusion by concatenating different imaging (e.g., PET, MRI) as well as other (e.g., demographics) modalities to form a single feature matrix to feed the model, resulting in conflated and non-specific analysis. As an illustration here, the PET and MRI data are both in the tabular form consisting of a hierarchy of regional summary statistics, while the demographic variates like age, sex, etc. are also stored in a table. SUVR stands for Standardized Uptake Value Ratio.
  \textbf{B.} MREF-AD introduces multimodal, regional experts design through MoE, ensuring the subsequent importance analysis are at brain region/group level rather than per feature column. More technical details can be seen later in \textbf{Fig.~\ref{fig:moe-ad-schematic}}. 
  \textbf{C.} As a result, MREF-AD enables brain region-aligned, subject-specific explainability. Here, two subjects with early AD (high risk) and stable MCI (lower risk) are shown as examples.}
  \label{fig_motiv}
  \vspace{-15pt}
\end{figure*}

Despite their utility, current multimodal models often rely on simple feature concatenation when handling structured tabular data, where features across modalities are analyzed as a single input and treated as equally informative \cite{liu2014multimodal, suk2015latent} (\textbf{Fig.~\ref{fig_motiv}}). Hence, this design ignores how the predictive value of different brain regions across different modalities might vary from patient to patient. Furthermore, interpreting these tabular models is difficult. Standard methods offer interpretability at the level of individual columns, or specific ROIs. For clinicians, the granularity of per-ROI analysis that treats localized anatomical points in isolation is often too fragmented to be insightful. Consequently, there is a lack of models that can provide mesoscopic brain regional interpretations, where the individual point ROI measurements are grouped into broader, clinically relevant brain regions such as the frontal or temporal lobes, that clinicians could use to evaluate disease progression and pathology \cite{alsaleh2025interpretable-multimodal-deep-learning-for-ad-diagnosis-lime-imaging-clinically-relevant-brain-regions, jagust2024adni-pet-core-at-20}.

Furthermore, in real-world clinical settings, multimodal data are frequently incomplete for some or all patients due to variability in imaging resources, institutional constraints, and patient-specific factors. While structural MRI is currently widely accessible, other more costly and specialized technologies, such as PET, remain limited in availability \cite{rabinovici2025updated-appropriate-use-criteria-for-amyloid-and-tau-pet}. Moreover, certain patients may be unable to undergo specific imaging procedures due to contraindications, including biomedical implants, pre-existing comorbidities, or physical and behavioral limitations that prevent the acquisition of a complete multimodal dataset \cite{dill2008mri-contraindications, johnson2013amyloid-contraindications}. Yet, standard multimodal frameworks typically deteriorate in performance in the event of missing modality \cite{ma2022are-multimodal-transformers-robust-to-missing-modality, wu2024deep-multimodal-learning-with-missing-modality-a-survey}. As such, they often require complete data, which excludes a significant portion of the broader patient population from model-assisted diagnosis \cite{abdelaziz2021alzheimer-disease-diagnosis-framework-from-incomplete-multimodal-data-using-cnn}. Thus, ensuring robust model performance despite missing modality at inference time is crucial for clinical deployment \cite{stahlschmidt2022multimodal-deep-learning-for-biomedical-data-fusion-review}. This avoids the impractical need of maintaining separate models for different data or modality combinations to serve all patients when new cohorts of data become available.

To address these limitations, we propose \textbf{MREF-AD}, a \emph{Multimodal Regional Expert Fusion} framework for AD diagnosis built upon an MoE design. MREF-AD decomposes each tabular neuroimaging modality—amyloid PET and structural MRI—into mesoscopic regional experts (e.g., frontal, temporal, parietal lobes), where each expert analyzes corresponding local imaging features (e.g., cortical thickness or regional SUVR values; \textbf{Fig.~\ref{fig:moe-ad-schematic}}). The gating mechanism enables MREF-AD to capture both cross-modality and within-modality (regional) heterogeneity, facilitating subject-specific and region-level interpretability. Each expert is implemented as a lightweight three-layer multilayer perceptron (MLP), ensuring that observed improvements arise from adaptive fusion rather than increased capacity. Furthermore, the gating network learns subject-specific fusion weights, dynamically emphasizing modality- and region-specific information most relevant to each individual.

We evaluated MREF-AD on data derived from the ADNI database, where the task is to classify subjects into cognitively normal (CN), mild cognitive impairment (MCI), and AD groups. The experimental results show that MREF-AD outperforms other baselines trained on concatenated imaging and demographic features, achieving improved accuracy, F1 score, and AUROC while offering interpretable insights into how brain region-specific structural and molecular imaging jointly contribute to disease classification. These results demonstrate the promise of our proposed adaptive and interpretable multimodal fusion for Alzheimer's disease diagnosis.

Our contributions can be summarized as follows:
\begin{enumerate}
    \item We introduce MREF-AD, an MoE architecture that decomposes each modality into regional experts and learns subject-specific modality--region fusion weights for adaptive, interpretable multimodal imaging.
    \item We show that this regional expert fusion improves three-way AD classification on the ADNI cohort compared to a set of representative baselines, especially in the event of missing modality.
    \item We provide region-level interpretability analyses that reveal how structural MRI and amyloid PET features are differentially prioritized across brain regions, yielding a data-driven atlas of regional biomarker relevance.
\end{enumerate}

\section{Related Work}

% \textcolor{red}{should we mention for interpretability, other models can do per column feature importance inherently, or they have to incorporate SHAP, and theres no flexibility to directly do groups of columns that may be more clinically relevant?
% remove GNN, add tabPFN?}

Early approaches to AD diagnosis classification often rely on straightforward multimodal fusion such as feature concatenation or kernel-based combinations~\cite{liu2014multimodal, suk2015latent, suk2014hierarchical, zhang2011multimodal, venugopalan2021multimodal,yi2023xgboost}.
Many machine learning and deep learning-based studies have adopted early fusion strategies, combining modality-specific features into a single, flattened representation, followed by some downstream classifiers~\cite{liu2014multimodal, suk2015latent, zhang2011multimodal}.
Other studies explored late fusion, for example, through ensemble learning or hierarchical fusion of single-modality predictors~\cite{suk2014hierarchical, venugopalan2021multimodal, yi2023xgboost}.
Among the various methods, ensemble-based Random Forest (RF) and XGBoost have demonstrated robust performance on AD prediction by capturing complex, non-linear interactions between multimodal features~\cite{bao2024combined, yi2023xgboost, guo2025machine}, while Logistic Regression (LR) remains a clinical standard due to its simplicity and clarity~\cite{li2025prediction}.
However, simple early fusion assumes all modalities are present and equally relevant and late fusion typically ignores the complex interactions across modalities. They provide limited insight into which modality or brain region drives a given prediction and often generate suboptimal results.

Modern approaches pursue more adaptive and interpretable multimodal fusion, leveraging techniques such as attention mechanisms, graph neural networks, contrastive learning, etc.~\cite{liu2018multi, meng2023research, klepl2022eeg, zhou2022interpretable, zhou2024multi, li2022multi, bahdanau2014neural, vaswani2017attention, tsai2019multimodal, hollmann2022tabpfn, hollmann2025accurate, gorishniy2021revisiting}
% For instance, graph neural networks have been introduced to model brain-region connectivity, yielding more biologically grounded representations and highlighting important network nodes in AD classification~\cite{meng2023research, klepl2022eeg, zhou2024multi, li2022multi, zhou2022interpretable}.
In particular, Transformer-derived attention mechanisms have emerged as the dominant strategy for integrating heterogeneous modalities. 
They enable learned feature weighting across modalities, improving interaction between multimodal signals~\cite{bahdanau2014neural, vaswani2017attention, tsai2019multimodal}.
As an example, FT-Transformer~\cite{gorishniy2021revisiting} represents one of the most significant works adapting the Transformer architecture for structured tabular data.
It tokenizes each feature into an embedding and feeds all embeddings to Transformer layers with self-attention to capture high-order, non-linear feature interactions.
However, FT-Transformer still takes multimodal input as a flat concatenation of features and cannot explicitly encode modality/brain region contributions.
Moreover, attention-based methods in general are sensitive to missing modality and require large training data with heavy computations due to the model complexity.

In addition, interpreting these methods for AD prediction remains challenging.
Conventional classifiers such as Logistic Regression, Random Forest, XGBoost, and MLPs usually provide feature importance either natively (from the coefficients or split-based importance in tree ensembles) or through post-hoc explainability analyses (most notably SHAP~\cite{lundberg2017unified}).
Although these approaches can effectively rank individual features such as neuroimaging traits and methods like SHAP have been increasingly adopted to explain complex black-box models for AD~\cite{yi2023xgboost, vahid2025predicting}, they generally operate at the granularity of individual features and provide per-column feature importance of the input data matrix.
Aggregating these dispersed feature attributions into clinically meaningful insights (e.g., temporal lobe regional contribution) requires manual post-processing and may not faithfully reflect how the models utilize multi-modal (e.g., MRI and PET neuroimaging), grouping signals (e.g., brain structures/regions) during prediction.

Recently, methods based on mixture-of-experts strategies have emerged to achieve adaptive fusion and emphasize intrinsic interpretability by the flexible experts design for several biomedical tasks including AD~\cite{yun2024flexmoe, jiang2025m, xin2025i2moe, ding2025denseformer, jiang2024m4oe, li2025m4}.
MoE architectures use a gating network to select among multiple expert subnetworks, allowing the model to tailor the feature processing to each input subject.
Yun et al. developed Flex-MoE for AD classification, which implemented generalized and specialized routers through a sparse MoE structure to deal with missing modality within imaging, clinical, biospecimen, and genetic multimodal data~\cite{yun2024flexmoe}. 
In another work, Jiang et al. proposed a multi-task multi-gate MoE framework, M$^3$AD, for simultaneous AD diagnosis and cognitive decline prediction from structural MRI images ~\cite{jiang2025m}.
It utilizes specialized experts for diagnosis-specific pathological patterns and shared experts for common structural features across cognitive decline progression.
Despite their strong performance across heterogeneous datasets, current MoE models for AD only operate at the whole-modality level and are not able to detect region-specific contributions within each modality.

% Previous multimodal approaches have used concatenation \cite{liu2014multimodal, suk2015latent}, convolutional fusion \cite{liu2018multi}, or ensemble and graph-based methods \cite{meng2023research, klepl2022eeg, zhou2024multi, li2022multi}, but they typically rely on fixed or global weighting schemes. Attention- and transformer-based fusion models \cite{bahdanau2014neural, vaswani2017attention, tsai2019multimodal} improve feature interaction yet remain sensitive to missing data. Recent mixture-of-experts (MoE) frameworks, such as Flex-MoE \cite{yun2024flexmoe}, have been proposed to dynamically weight multiple modalities, achieving strong performance across heterogeneous datasets. However, these models operate at the whole-modality level and do not provide insight into region-specific contributions within each modality. 

\section{Methods}
\label{sec:methods}

Data used in the preparation of this paper were obtained from the ADNI database\footnote{\url{http://adni.loni.usc.edu}}. The ADNI was launched in 2003 as a public-private partnership led by Principal Investigator Michael W. Weiner, MD. The primary goal of ADNI has been to test whether serial MRI, PET, other biological markers, and clinical and neuropsychological assessment can be combined to measure the progression of mild cognitive impairment (MCI) and early AD. All participants provided written informed consent, and study protocols were approved by each participating site’s Institutional Review Board (IRB). Up-to-date information about the ADNI is available at \url{www.adni-info.org}.

This study focuses on the multimodal neuroimaging data amyloid and MRI subset of the ADNI \cite{weiner2017recent} dataset, as well as demographic information (age, sex, education, race, and ethnicity). Amyloid PET and MRI were selected as representative molecular and structural biomarkers, providing complementary information on amyloid-$\beta$ deposition and neurodegenerative atrophy, respectively.

After preprocessing and quality control, the final dataset included 1,530 unique participants, each contributing one imaging visit corresponding to their last available amyloid–MRI session (Table~\ref{tab:participant_info}). Each visit was labeled with the participant's clinical diagnosis of cognitively normal (CN), mild cognitive impairment (MCI), or Alzheimer's disease (AD).

Amyloid PET features were aligned with MRI features by matching both subject ID and visit code to ensure same-session correspondence between modalities. For each modality, region-level measurements were defined using the FreeSurfer cortical and subcortical parcellation atlas~\cite{fischl2012freesurfer, desikan2006automated}. By mapping individual brain scans into a standardized set of ROI measures, FreeSurfer derives morphometric features (68 in total from left and right hemispheres), which are widely utilized as a testbed for evaluating computational models within AD research and serve as a robust and reproducible benchmark. For amyloid PET, we extracted standardized uptake value ratios (SUVRs) indicating regional amyloid-$\beta$ plaque burden, and for MRI, regional volumes (mm³) of the brain. 

\begin{table}[t]
\centering
\caption{Participant demographics at the last available visit in the ADNI dataset. Values are reported as mean~$\pm$~standard deviation unless otherwise noted. Participant counts refer to unique individuals, with diagnoses assigned based on their last recorded imaging session.}
% \vspace{5pt}
\resizebox{\columnwidth}{!}{
\begin{tabular}{lcccc}
\hline
\textbf{Characteristic} & \textbf{CN} & \textbf{MCI} & \textbf{AD} & \textbf{Total} \\
\hline
Participants (n) & 637 & 557 & 336 & 1530 \\
Age (years) & 74.8~$\pm$~8.0 & 75.2~$\pm$~8.0 & 76.6~$\pm$~8.2 & 75.3~$\pm$~8.1 \\
Sex (M/F) & 268/369 & 304/253 & 192/144 & 764/766 \\
Education (years) & 16.7~$\pm$~2.4 & 16.0~$\pm$~2.7 & 15.9~$\pm$~2.6 & 16.3~$\pm$~2.6 \\
\hline
\end{tabular}
}
\label{tab:participant_info}
% \vspace{-15pt}
\end{table}

\subsection{Model Architecture}

We formulate the three-way Alzheimer’s disease classification task as a multimodal learning problem. Let $\mathbf{x} = \{ \mathbf{x}_1, \dots, \mathbf{x}_N \}$ represent input features from $N$ modality-region pairs. Our objective is to learn a robust mapping that predicts the diagnostic label $y \in \{\mbox{CN}, \mbox{MCI}, \mbox{AD}\}$ even when specific features are missing.

\begin{figure}[tbp]
    \centering
    \includegraphics[width=0.9\linewidth] {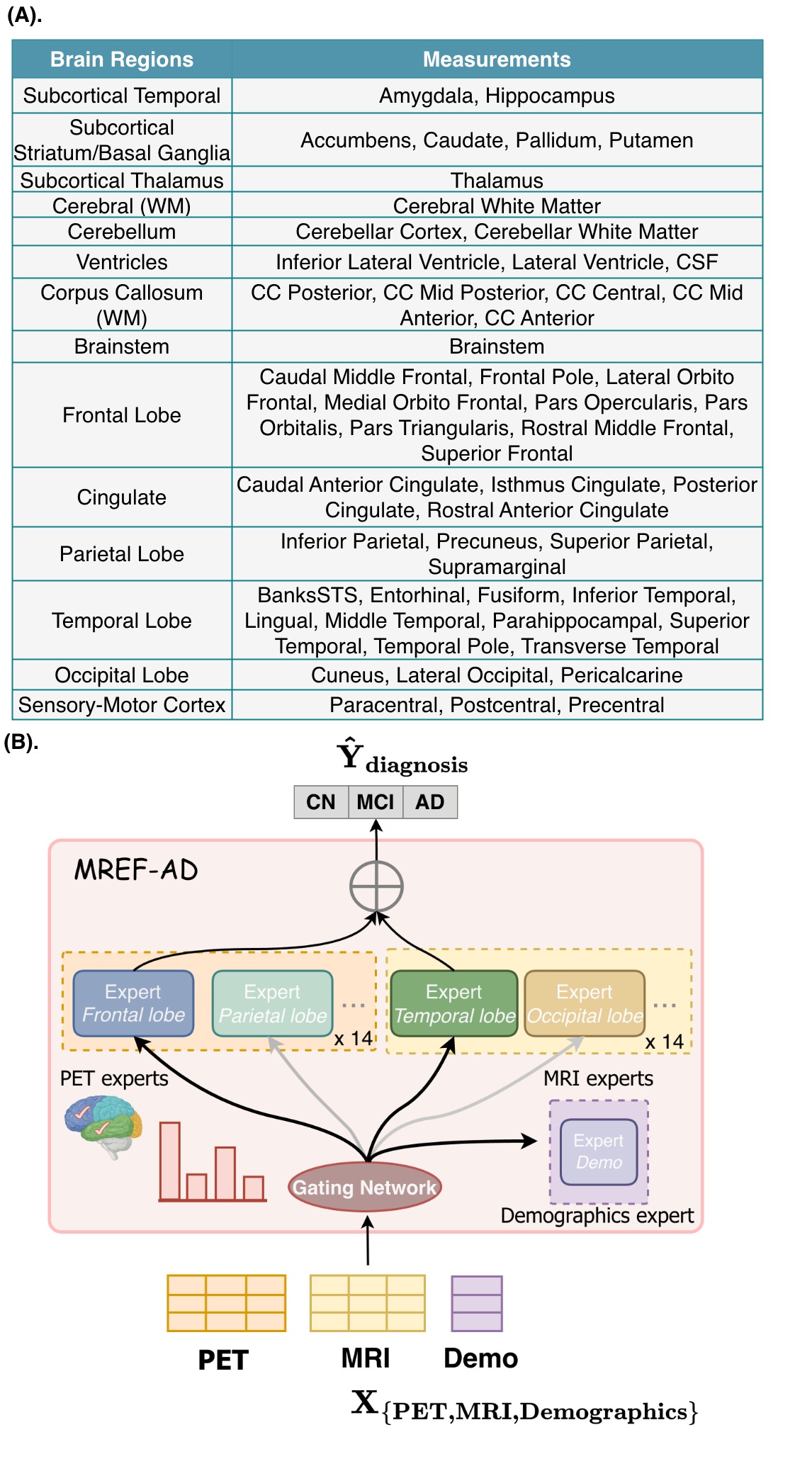}
    % {Figures/MREF-AD_team-fig1_v6.png}
        % \includegraphics[width=1\linewidth]{Figures/MREF-AD_schematic_fig2_v6.pdf}
    \vspace{-15pt}
    \caption{\textbf{MREF-AD framework.}
    (A) Brain regions (n = 14) in the neuroimaging data and their corresponding FreeSurfer measurements.
    (B) In MREF-AD, 28 region-level experts from amyloid PET and MRI, together with one demographic expert, model modality–region features. Gating network assigns expert weights for Alzheimer's disease diagnosis, enabling brain region-level interpretability.}
    \label{fig:moe-ad-schematic}
    \vspace{-15pt}
\end{figure}

MREF-AD (\textbf{Fig.~\ref{fig:moe-ad-schematic}}) is an MoE model of $N$ modality--region experts and a learned gating network that adaptively fuses their outputs. In our implementation, each neuroimaging modality (amyloid PET and MRI) is partitioned into 14 mesoscopic brain regions, yielding 28 imaging experts. Furthermore, we include demographic variables and treat them as an additional expert within our multimodal fusion framework. Thus, this results in a total of $N=29$ experts (28 imaging experts + 1 demographic). %Specifically, each expert $f_m(\mathbf{x}_m)$ is a three-layer multilayer perceptron (MLP) that maps region-specific input features to class logits $\mathbf{h}_m \in \mathbb{R}^{C}$ for $C=3$.

\textbf{Expert Networks.} Each expert $f_m(\mathbf{x}_m)$ is implemented as a three-layer multilayer perceptron (MLP) with architecture [$\mathbf{d}_m$,H1,H2,C], where $\mathbf{d}_m$ is the input dimension for expert (varying by modality-region pair) , H1 and H2 are hidden dimensions, and C is the number of output classes.  Each expert maps region-specific input features to class logits $\mathbf{h}_m \in \mathbb{R}^{C}$ for $C=3$. 
% \textcolor{red}{[Moved the rest commented out to the Final Prediction]}
% The gating network produces logits $\mathbf{z} \in \mathbb{R}^N$, which are converted to non-negative fusion weights $g = \mathrm{softmax}(\mathbf{z})$ with $\sum_{m=1}^N g_m = 1$. The final prediction is
% \begin{equation}
% p(y \mid \mathbf{x}) = \mathrm{softmax}\!\left(\sum_{m=1}^{N} g_m \mathbf{h}_m\right),      \label{eq1}
% \end{equation}
% where $y$ is the diagnostic class (CN, MCI, AD). Missing modalities are handled by masking their experts before renormalizing the gate.

%Original commented out
% \textbf{Hierarchical Gating Network.}
% To capture within-modality structure, MREF-AD employs a two-level hierarchical gating scheme. A modality-level gate first allocates weights across imaging and demographic modalities (MRI, PET, demographics), and a regional gate within each modality then distributes the modality weight among its regional experts. The final expert weight is proportional to the product of its modality and region gates, providing subject-specific modality--region weights and modeling both cross-modality and within-modality heterogeneity in a unified framework. Specifically, the mixture weights $g_m$ are produced by a hierarchical gating network. Internally, the gate first computes modality-level weights $w^{\mathrm{mod}}_{i,k}$ that reflect the relative importance of each modality for subject $i$, and then computes region-level weights $w^{\mathrm{reg}}_{i,m}$ that distribute each modality’s weight across its regional experts. 
\textbf{Gating Network.} To capture within- and cross-modality structure, MREF-AD employs a gating scheme that produces subject-specific fusion weights. The gating network first computes modality-level logits $\mathbf{z}^{\mathrm{mod}} \in \mathbb{R}^{K}$ for $K$ modalities (MRI, PET, demographics), which are converted to modality weights via:

\begin{equation}
w^{\mathrm{mod}}_{i,k} = \frac{\exp(z^{\mathrm{mod}}_{i,k} / \tau)}{\sum_{k'=1}^{K} \exp(z^{\mathrm{mod}}_{i,k'} / \tau)} .
\label{eq:modality_weights}
\end{equation}
where $\tau$ is a temperature hyperparameter that controls the confidence of the gating distribution. A lower temperature (e.g., $\tau \to 0$) pushes the weights toward a one-hot distribution, effectively selecting the single most relevant expert, while a higher temperature encourages a more uniform contribution across experts.

For each modality $k$, the gating network then computes region-level logits $\mathbf{z}^{\mathrm{reg}}_k \in \mathbb{R}^{R_k}$ for the $R_k$ regions within that modality, which are normalized to regional weights:

\begin{equation}
w^{\mathrm{reg}}_{i,m} = \frac{\exp(z^{\mathrm{reg}}_{i,m} / \tau)}{\sum_{m' \in \mathcal{R}_k} \exp(z^{\mathrm{reg}}_{i,m'} / \tau)} ,
\label{eq:region_weights}
\end{equation}

where $\mathcal{R}_k$ denotes the set of regional experts within modality $k$. 
To promote expert diversity and prevent the gating network from converging to a sub-optimal subset of experts, we add Gaussian noise to the gating weights $w$ prior to the softmax operation:
\begin{equation}
\tilde{w} = w + \sigma_{\text{noise}} \cdot \epsilon, \quad \epsilon \sim \mathcal{N}(0, I)
\label{eq:gate_noise}
\end{equation}
where $\sigma_{\text{noise}}$ is a scalar hyperparameter representing the noise scale. For experts corresponding to missing modalities, the noisy weights $\tilde{w}$ are masked to $-\infty$ before applying the temperature-scaled softmax: $\text{softmax}(\tilde{w}/\tau)$.
The final expert weight combines these two levels:
\begin{equation}
g_{i,m} = w^{\mathrm{mod}}_{i,k(m)} \cdot w^{\mathrm{reg}}_{i,m},
\label{eq:gate_composition}
\end{equation}
where $k(m)$ denotes the modality associated with expert $m$. Note that $\sum_{m=1}^N g_{i,m} = 1$ by construction. 
While MREF-AD is designed to support this hierarchical gating scheme, the framework also allows for a flat gating configuration where a single gating network assigns weights across all 29 modality-region experts simultaneously. For the primary results reported in this study, the flat gating configuration was utilized as it was empirically found to offer superior diagnostic performance and more flexible weight allocation.

%Original commented out
% \textbf{Final Prediction.} The final expert weight is obtained by combining these two levels and renormalizing:
% \begin{equation}
% g_{i,m}
% \;=\;
% \frac{
% w^{\mathrm{mod}}_{i,k(m)} \, w^{\mathrm{reg}}_{i,m}
% }{
% \sum_{m'} w^{\mathrm{mod}}_{i,k(m')} \, w^{\mathrm{reg}}_{i,m'}
% },
% \label{eq:gate_composition}
% \end{equation}
% where $k(m)$ denotes the modality associated with expert $m$.Although we focus on amyloid PET, MRI, and demographics in this study, the formulation is modality-agnostic and can be directly applied to other multimodal imaging settings.

To regularize the gating mechanism, we added a sparsity penalty based on the entropy of the gating weights and a diversity penalty encouraging decorrelated expert outputs. The total loss function is:
\begin{equation}
\mathcal{L} = \mathcal{L}_{\text{CE}} + \lambda_1 \mathcal{L}_{\text{sparsity}} + \lambda_2 \mathcal{L}_{\text{diversity}},
\label{eq:total_loss}
\end{equation}
where $\mathcal{L}_{\text{CE}}$ is the cross-entropy  with class weights inversely proportional to the training set frequencies to address class imbalance. The sparsity penalty minimizes the entropy of gating weights to encourage focused expert selection:
\begin{equation}
\mathcal{L}_{\text{sparsity}} = -\frac{1}{B}\sum_{i=1}^{B} \sum_{m=1}^{N} g_{i,m} \log g_{i,m},
\label{eq:sparsity}
\end{equation}
where $B$ is the batch size. The diversity penalty encourages decorrelated expert predictions by minimizing pairwise cosine similarities:
\begin{equation}
\mathcal{L}_{\text{diversity}} = \sum_{m=1}^{N} \sum_{m' > m} \frac{\mathbf{h}_m^T \mathbf{h}_{m'}}{\|\mathbf{h}_m\| \|\mathbf{h}_{m'}\|}.
\label{eq:diversity}
\end{equation}
Optimization used the AdamW optimizer (weight decay $1 \times 10^{-4}$) for up to 40 epochs, with early stopping based on validation loss (patience = 10). Hyperparameters were optimized using Optuna \cite{akiba2019optuna} with Tree-structured Parzen Estimator (TPE) sampling \cite{watanabe2023tpe} . The search space included learning rate ($10^{-5}$--$10^{-2}$, log-uniform), weight decay ($10^{-6}$--$10^{-3}$, log-uniform), dropout rate ($0$--$0.5$), and batch size $\{16, 32, 64, 128\}$. For the MREF-AD architecture, we searched expert hidden widths ($64$--$512$) and gating hidden widths ($32$--$256$). To stabilize expert selection, we utilized a temperature decay schedule $\tau_t = \max(\tau_{\min}, \tau_{\text{start}} \cdot \tau_{\text{decay}}^t)$, searching $\tau_{\min}$ ($0.05$--$1.0$), $\tau_{\text{start}}$ ($0.5$--$1.5$), and $\tau_{\text{decay}}$ ($0.90$--$0.999$), alongside a gate noise factor $\sigma$ ($0$--$0.1$). The best-performing configuration utilized a learning rate of $4.25 \times 10^{-4}$, weight decay of $1.17 \times 10^{-5}$, dropout of $0.04$, and a batch size of $64$. The optimal architecture consisted of regional experts with a hidden dimension of $233$ and a gating network hidden dimension of $68$. This configuration employed a noise factor $\sigma \approx 0.015$ and a temperature schedule with $\tau_{\min}=0.56$, $\tau_{\text{start}}=0.71$, and $\tau_{\text{decay}}=0.907$. Regularization was enforced via an entropy-based sparsity penalty $\lambda_1 \approx 0.11$ (searched $0$--$0.2$) and a diversity penalty $\lambda_2 \approx 0.06$ (searched $0$--$0.1$).

\textbf{Final Prediction.} The model's output combines all expert predictions weighted by their gating scores:
\begin{equation}
p(y \mid \mathbf{x}) = \mathrm{softmax}\!\left(\sum_{m=1}^{N} g_{i,m} \mathbf{h}_m\right),
\label{eq:final_prediction}
\end{equation}
where $y \in \{\text{CN}, \text{MCI}, \text{AD}\}$. Missing modalities are handled by masking their corresponding experts (setting $z^{\mathrm{mod}}_{i,k} = -\infty$ for missing modality $k$) before applying the softmax normalization, which naturally excludes them from the weighted sum.

Although we focus on amyloid PET, MRI, and demographics in this study, the formulation is modality-agnostic and can be directly applied to other multimodal imaging settings.

\subsection{Handling Missing Modality}
In real-world clinical practice, multimodal data are often incomplete and modalities might be missing at inference time. To evaluate model robustness, we simulated clinical scenarios where a modality is unavailable by withholding all features for a specific modality (e.g., amyloid PET or MRI). Specifically, all models are trained on complete data, while at inference time, a modality is nulled by removing the numerical values within the test sets.

MREF-AD is explicitly designed to handle missing modalities by (i) masking and excluding missing modalities or experts from the gating distribution and (ii) normalizes the remaining gate weights such that predictions are based only from observed inputs.

To ensure that every expert in MREF-AD receives a valid numerical input even when a modality is unavailable, we first impute missing values with the training set median. Binary availability indicators $\mathds{1} \in \{0, 1\}$ are associated with each expert and propagated to the gating network. These indicators mask expert participation at the gating stage and mathematically guarantees that missing experts receive zero mixture weight. Formally, let $\mathds{1}_{i,k}^{\text{mod}}$ be the availability of modality $k$ and $\mathds{1}_{i,m}^{\text{reg}}$ be the availability of region $m$ for subject $i$. The gating weights $w^{\text{mod}}$ and $w^{\text{reg}}$ are constrained such that:
\begin{equation}
\mathds{1}_{i,k}^{\text{mod}} = 0 \implies w_{i,k}^{\text{mod}} = 0, \quad \mathds{1}_{i,m}^{\text{reg}} = 0 \implies w_{i,m}^{\text{reg}} = 0
\end{equation}
Hence, the masking ensures that the final prediction $p(y|\mathbf{x})$ is computed exclusively from available data.

As such, MREF-AD is able to perform inference on missing modality through modality and expert masking, without the need for retraining or modality-specific model variants. Furthermore, the learned expert weights provide a sample-specific explanation of expert contribution to each prediction, conditional on availability.

Baseline models (detailed below) that cannot natively process missing inputs similarly apply median imputation, or mean statistics  following their default implementation.

\subsection{Baselines}

To provide fair comparisons, we first implemented a multimodal fusion baseline using the same three-layer MLP building block as MREF-AD. Furthermore, we compared our approach against state-of-the-art architectures for tabular and multimodal data. First, we compared with Flex-MoE~\cite{yun2024flexmoe}, a sparse MoE framework specifically designed for AD classification with multimodal inputs. We also compared against the sophisticated tabular-attention model, FT-Transformer~\cite{gorishniy2021revisiting}, other tabular state-of-the-art models TabNet~\cite{alsaleh2025interpretable-multimodal-deep-learning-for-ad-diagnosis-lime-imaging-clinically-relevant-brain-regions}, TabM~\cite{gorishniy2024tabm}, and NODE\cite{popov2019NODE}, as well as traditional machine learning classifiers (Random Forest, Logistic Regression, XGBoost) . Except for Flex-MoE, which requires modality specifications, all baselines receive the same concatenated features across modalities and regions.

To ensure rigorous comparison, MLP and FT-Transformer were trained with cross-entropy loss and class-balanced weights to address label imbalance, matching that of MREF-AD. Other neural baselines follow unweighted training in the main results due to an ablation showing that adding inverse-frequency weights did not improve their performance. 

% This combined block ensures Table II is on top and Figure 3 is immediately below it.
\begin{table}[t!]
\centering
\caption{Diagnostic performance comparison between traditional and deep multimodal fusion methods across three classes of AD diagnosis (CN, MCI, AD). Results are reported as mean $\pm$ standard deviation across ten seeds.}
\resizebox{\columnwidth}{!}{%
\begin{tabular}{lcc}
\hline
\textbf{Model} & \textbf{Accuracy} & \textbf{F1-score} \\ 
\hline
Logistic Regression & $0.636 \pm 0.019$ & $0.628 \pm 0.019$ \\
Random Forest & $0.638 \pm 0.030$ & $0.635 \pm 0.032$ \\
XGBoost & $0.619 \pm 0.039$ & $0.613 \pm 0.038$ \\
\hline
MLP & $0.619 \pm 0.020$ & $0.608 \pm 0.024$ \\
FT-Transformer & $0.611 \pm 0.027$ & $0.598 \pm 0.027$ \\
NODE & $0.644 \pm 0.024$ & $0.648 \pm 0.025$ \\
TabNet & $0.593 \pm 0.050$ & $0.578 \pm 0.061$ \\
TabM & $0.599 \pm 0.034$ & $0.611 \pm 0.030$ \\
Flex-MoE & $0.597 \pm 0.033$ & $0.584 \pm 0.037$ \\ 
\textbf{MREF-AD (ours)} & \textbf{0.657 $\pm$ 0.024} & \textbf{0.664 $\pm$ 0.028} \\
\hline
\end{tabular}%
}
\label{tab:moe_results_10cv_lastvisit}

\vspace{2em} % Adds a clean gap between Table and Figure

\includegraphics[width=\linewidth]{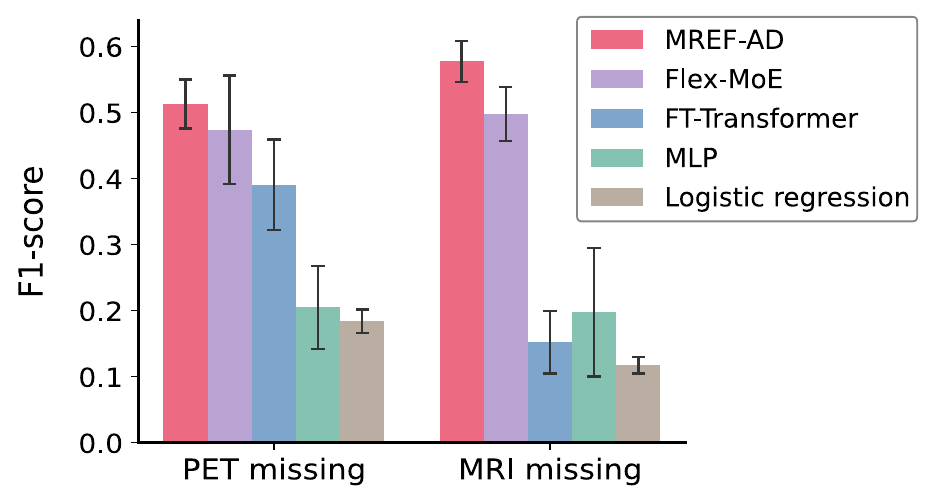}
\captionof{figure}{\textbf{Model robustness to missing modalities at inference time.} Comparison of F1-scores for MREF-AD and baselines under missing amyloid PET or MRI data at inference time. Error bars indicate standard deviation across 10 seeds.}
\vspace{-10pt}
\label{fig:missing_modality}
\end{table}

\vspace{-5pt}

\subsection{Training and evaluation}

We evaluated all models by subsampling across 10 random seeds, partitioning into training, validation, and test sets with a 80/10/10 ratio, respectively. For each seed, we performed hyperparameter optimization for each model with Optuna \cite{akiba2019optuna} on the training and validation subsets. To ensure sufficient search space exploration of hyperparameters, we ran 100 Optuna trials for the traditional machine learning baselines and 200 trials for neural network-based architectures, including MREF-AD. We then obtained the best-performing hyperparameters for each model, which were used to train the final model on the combined training and validation data. The model's performance was evaluated on the held-out test set. Finally, we report the accuracy and macro-F1 mean and standard deviation across all 10 seeds to provide a robust assessment of model performance and stability. All features were z-score normalized within each training fold and the corresponding statistics were applied to the validation data.

% \vspace {-5pt}

\section{Results}
\label{sec:results}

Table~\ref{tab:moe_results_10cv_lastvisit} summarizes performance on the three-way diagnostic task (CN vs MCI vs AD). Under optimal conditions with complete data modality, MREF-AD consistently outperforms the baseline and state-of-the-art models. While traditional ML models seemingly provide strong baselines, MREF-AD achieves significant improvements in F1 score over them and neural baselines  (paired one-sided Wilcoxon tests with Holm correction, p $<$ 0.02), including FT-Transformer, TabNet, TabM, and Flex-MoE. However, while traditional and neural models provide competitive performance in these optimal settings, they lack the architectural flexibility to handle missing data, that is common in clinical practice. 

\vspace{-5pt}

\subsection{Robustness to Missing Modalities}

Next, we evaluate the robustness of MREF-AD against representative baselines under realistic clinical settings, where one modality is missing at inference. 

Taking models trained on the complete dataset and removing one of the neuroimaging modalities from the test data at inference time, MREF-AD outperforms all baselines in both missing PET and MRI modality conditions (\textbf{Fig.~\ref{fig:missing_modality}}). In both missing modality conditions, MREF-AD maintains an F1-score of above $0.50$. The next best-performing baselines are Flex-MoE, which is designed to handle missing modality, with F1-score struggling to pass 0.50, followed by FT-Transformer, with an F1-score that drops to approximately $0.38$. Interestingly, FT-Transformer's performance decreases significantly in the event of missing MRI, suggesting that it is overly reliant on MRI modality. Finally, MLP and Logistic Regression yield significantly lower performance near $0.20$ in both missing modality conditions. 

Moreover, although Logistic Regression achieves competitive performance under full-modality conditions 
%that surpass other baseline models 
(Table \ref{tab:moe_results_10cv_lastvisit}), 
notably, it shows the most significant performance decline in the event of missing modality. In contrast, MREF-AD demonstrates robustness to missing modality inputs. This shows that the handling of missing modality within its architecture helped the model maintain the highest diagnostic performance.

\subsection{Clinical Interpretability}
\begin{figure}[t!]
    \centering
     \includegraphics[width=0.8\linewidth]{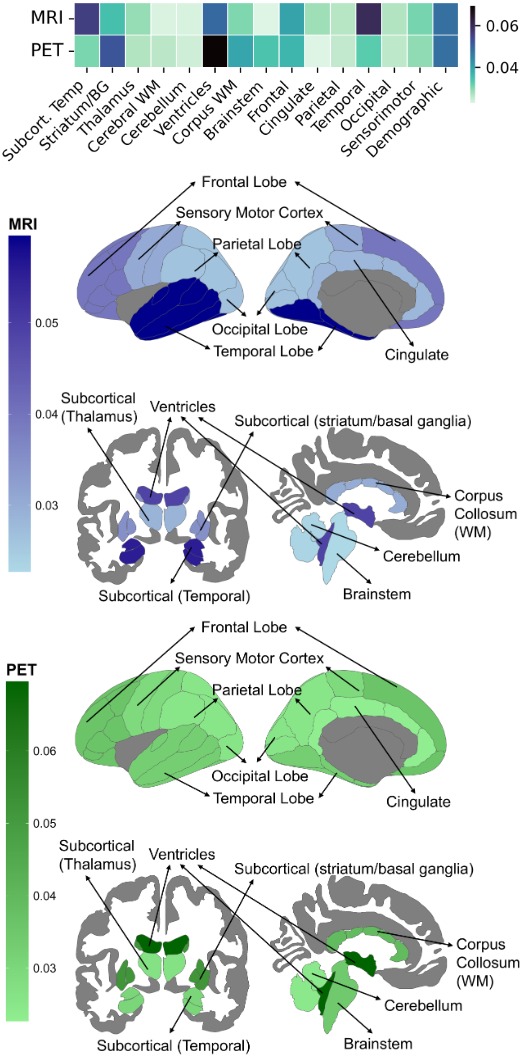}
    \vspace{-5pt}
    \caption{\textbf{Average modality–region expert contributions.} Top: mean gating weights across subjects for each regional expert, grouped by modality (MRI, PET, demographics). Middle and bottom: MRI and amyloid PET expert contributions projected onto cortical and subcortical regions; Color intensity (darker shades) indicates greater model reliance on that regional expert.}
    \label{fig:subject_heatmap}
    \vspace{-10pt}
\end{figure}

\begin{figure} [t!]
    \centering
    \includegraphics[width=\linewidth]{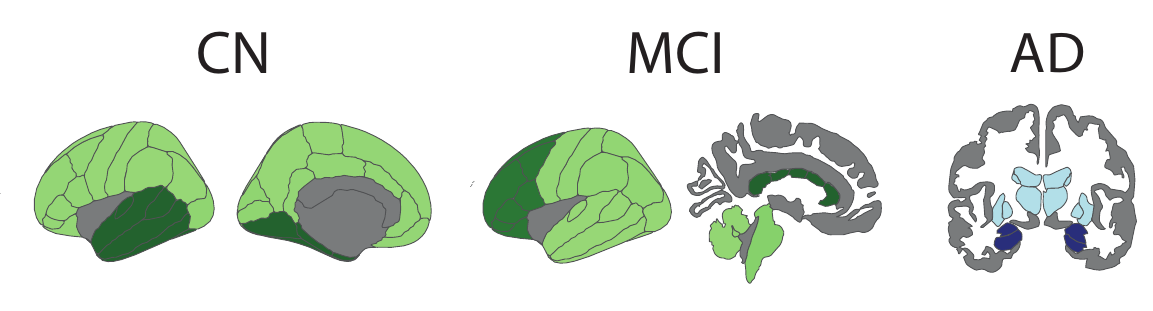}
    
    \vspace{-10pt}
    \caption{\textbf{Subject-specific expert contributions.} Comparison of gating weights for CN (left), MCI (middle), and AD (right) patients. The color intensity is shown on the same scale as in Fig.~\ref{fig:subject_heatmap}}
    
    \vspace{-10pt}
    \label{fig:subject_specific_viz}
\end{figure}

The gating network in MREF-AD produces sample-specific softmax weights over experts, determining their relative influence on the final prediction. This enables model interpretability by aggregating these weights across subjects to obtain expert contributions that quantify the average importance of each regional expert in diagnostic decision-making.

\textbf{Fig.~\ref{fig:subject_heatmap}} summarizes these contributions across test sets. Both MRI and amyloid PET experts receive substantial weights, with MRI experts showing slightly higher aggregate contributions. Among MRI experts, Temporal, Subcortical Temporal and Ventricular regions rank highly, suggesting that structural alterations in these areas offer complementary information for disease classification. On the amyloid side, regions such as Brainstem, Striatum/Basal Ganglia, and Corpus Callosum White Matter exhibit prominent contributions, reflecting strong molecular cues. Several regions,including the Subcortical Temporal area, show high contributions in both modalities, indicating convergent molecular and structural relevance, whereas regions such as the Brainstem exhibit dominant importance within a single modality, demonstrating the model's capacity to disentangle modality-specific signals. The Demographic expert ranks among the highest-contributing experts, underscoring the consistent influence of subject-level variables. 

The averaged pattern shows the model’s overall behavior, however MREF-AD also offers interpretability for individual clinical profiles. To illustrate this, 
\textbf{Fig.~\ref{fig:subject_specific_viz}} shows the model's subject-specific gating weights for representative CN, MCI, and AD cases, showing how the model shifts its emphasis across the disease stages. For a CN individual, the temporal lobe in PET receives the highest weight. In MCI, the model distributes more weight on the Corpus Callosum and frontal regions in PET. In an AD patient, the weighting shifts toward subcortical temporal structures in MRI.

% Together, the examples show that MREF-AD offers personalized interpretability into each patient's diagnosis and adapts its regional and modality-level focus in a manner aligned with known clinical progression of Alzheimer’s disease.

Overall, these examples show that MREF-AD adaptively balances shared and modality-specific regional information. The prioritization of temporal and subcortical regions are consistent with established AD biomarkers found through imaging \cite{jack2010hypothetical, frisoni2010clinical, jagust2018imaging}. Hence, we see that the learned expert weights yield a data-driven brain map with biomarker relevance that align with known clinical progression of AD, allowing for personalized interpretability into each patient's diagnosis.

\begin{table}[t!]
\centering
\caption{Ablation and sparsity analysis of MREF-AD. Values are mean~$\pm$~SD over 10 seeds.}
\resizebox{\columnwidth}{!}{
\begin{tabular}{lcc}
\hline
\textbf{Configuration} & \textbf{Accuracy} & \textbf{F1-score} \\
\hline
\textbf{MREF-AD} & \textbf{0.657 $\pm$ 0.024} & \textbf{0.664 $\pm$ 0.028} \\
\hline
\multicolumn{3}{l}{\emph{(A) Modality ablation}} \\
\quad w/o Amyloid & $0.592 \pm 0.023$ & $0.596 \pm 0.022$ \\
\quad w/o MRI & $0.610 \pm 0.021$ & $0.606 \pm 0.019$ \\
\quad w/o Demographic & $0.638 \pm 0.026$ & $0.644 \pm 0.030$ \\
\quad Amyloid only & $0.624 \pm 0.016$ & $0.615 \pm 0.023$ \\
\quad MRI only & $0.568 \pm 0.030$ & $0.570 \pm 0.034$ \\
\hline
\multicolumn{3}{l}{\emph{(B) Expert sparsity (top-$k$ gating)}} \\
\quad Top-5 experts & $0.620 \pm 0.025$ & $0.609 \pm 0.039$ \\
\quad Top-3 experts & $0.642 \pm 0.022$ & $0.643 \pm 0.030$ \\
\quad Top-1 expert  & $0.589 \pm 0.039$ & $0.562 \pm 0.039$ \\
\hline
\multicolumn{3}{l}{\emph{(C) Gating architecture ablation}} \\
\quad Modality-only gate & $0.653 \pm 0.029$ & $0.656 \pm 0.025$ \\
\quad Region and Modality gates & $0.634 \pm 0.015$ & $0.631 \pm 0.023$ \\
\hline
\end{tabular}
}
\label{tab:ablation}
\vspace{-12pt}
\end{table}

\vspace{-5pt}
\subsection{Ablations}
\vspace{-2pt}
Next, we conducted comprehensive ablation experiments to evaluate the robustness and design choices of MREF-AD (Table~\ref{tab:ablation}). Under leave-one-modality-out retraining settings, performance decreased modestly relative to the full model, indicating that MREF-AD effectively leverages complementary information across modalities. Among modalities, removing amyloid PET led to the largest performance decline while MRI removal produced smaller changes.

We further examined the effect of expert sparsity and gating design. Top-$k$ gating, specifically Top-3 sparse MREF-AD, yielded performance that still outperforms baselines. Hence, MREF-AD can achieve competitive performance with significantly fewer active experts, enhancing efficiency. While a hierarchical modality-region gating scheme was initially designed to capture structured dependencies, empirical results indicated that a flat regional gating architecture, of directly assigning weights across all 29 experts, offered the best diagnostic performance.

\vspace{-5pt}

\subsection{Model Complexity}

\vspace{-2pt}
% \begin{table}[tbp]
% \caption{Model Complexity and Computational Efficiency Comparison}
% \label{tab:complexity}
% \centering
% \begin{small}
% \resizebox{\columnwidth}{!}{
%     \begin{tabular}{@{}lrrr@{}}
%         \toprule
%         \textbf{Method} & \textbf{Active Params} & \textbf{Active FLOPs} & \textbf{Total Params} \\ \midrule
%         MLP (Baseline)           & 161,923          & 364,044          & 161,923 \\
%         FT-Transformer           & 941,379          & 2,666,760        & 941,379 \\ 
%         Flex-MoE & 629,891 & 1,424,550 & 1,948,291 \\
%         \midrule
%         \textbf{MREF-AD (Top-1)} & \textbf{111,544} & \textbf{221,820} & \multirow{4}{*}{456,092} \\
%         MREF-AD (Top-3)          & 135,036          & 268,364          &  \\
%         MREF-AD (Top-5)          & 158,528          & 314,908          &  \\
%         MREF-AD (Top-10) & 217,258 & 431,268 & \\ 
%         MREF-AD (Dense)          & 440,432          & 873,436          &  \\ \bottomrule
%     \end{tabular}
% }
% \end{small}
% \vspace{1mm}
% \begin{flushleft}
% \footnotesize{Note: Active parameters and FLOPs refer to the computational cost per single inference. Lower values indicate more efficient prediction. For MREF-AD, Top-$k$ denotes the number of experts activated by the gating network.}
% \end{flushleft}
% \vspace{-15pt}
% \end{table}

% \vspace{-3pt}
\begin{table}[tbp]
\caption{Model Complexity and Computational Efficiency Comparison}
\label{tab:complexity}
\centering
\begin{small}
\resizebox{\columnwidth}{!}{
    \begin{tabular}{@{}lrrr@{}}
        \toprule
        \textbf{Method} & \textbf{Active Params} & \textbf{Active FLOPs} & \textbf{Total Params} \\ \midrule
        MLP (Baseline) & 200,885 & 401,008 & 200,885 \\
        FT-Transformer & 1,000,515 & 2,666,760 & 1,000,515 \\
        Flex-MoE & 629,891 & 1,424,550 & 1,948,291 \\
        \midrule
        \textbf{MREF-AD (Top-1)} & \textbf{106,223} & \textbf{211,069} & \multirow{4}{*}{946,915} \\
        MREF-AD (Top-3) & 164,475 & 326,869 & \\
        MREF-AD (Top-5) & 222,727 & 442,669 & \\
        MREF-AD (Dense) & 921,751 & 1,832,269 & \\ \bottomrule
    \end{tabular}
}
\end{small}
\vspace{1mm}
\begin{flushleft}
\footnotesize{Note: Active parameters and FLOPs refer to the computational cost per single inference. Lower values indicate more efficient prediction. For MREF-AD, Top-$k$ denotes the number of experts activated by the gating network.}
\end{flushleft}
\vspace{-15pt}
\end{table}

While performance, robustness to missing modality, and clinical interpretability are crucial, the practicality of deep learning models in clinical settings is also influenced by the computational cost and ease of deployment \cite{abulibdeh2025balancing-model-complexity-and-clinical-deployability-in-deep-learning, sheikh2025lightweight}. As such, we evaluate the complexity of MREF-AD by comparing its total model parameters, the number of active parameters actively used during inference, and active Floating Point Operations (FLOPs) against representative neural network-based baselines (Table~\ref{tab:complexity}) based on their best hyperparameters. 

Our results show MREF-AD achieving parameter-efficient learning through its architectural design. It has a total model capacity of 946,915 parameters, smaller than the 1,000,515 parameters of the FT-Transformer and 1,948,291 for Flex-MoE. More importantly, while MREF-AD learns a global distribution over experts, it offers a tunable Top-$k$ inference mode. This allows for significant computational savings at deployment by only activating the $k$ most relevant experts identified by the gating network, whereas FT-Transformer and MLP activate all available parameters. In the Top-1 configuration, MREF-AD activates just 106,223 parameters per prediction, which significantly reduces the model's effective complexity at inference time. This translates to just 211,069 FLOPs per pass, which is roughly a 12.6-fold reduction compared to FT-Transformer (2.67M FLOPs) and makes MREF-AD even more efficient compared to the simpler MLP baseline (401K FLOPs). Similarly, in the Top-3 setting, the active parameters (164,475) and FLOPs (326,869) of MREF-AD remain lower than MLP while still maintaining a higher predictive performance (Tables~\ref{tab:moe_results_10cv_lastvisit} and \ref{tab:ablation}). In the dense setting, when all experts are activated, MREF-AD's complexity remains lower than FT-Transformer in terms of FLOPs, with 921,751 active parameters and 1.83M floating-point operations (vs. 2.67M). Ultimately, MREF-AD offers a flexible, tunable configuration for clinical deployment through the top-$k$ expert parameter, allowing the model to be adapted to different clinical environments with varying computational resources.

\vspace{-10pt}
\section{Discussion}
\label{sec:discussion}

% \textcolor{red}{todo: add more in the $\rightarrow$ done, please double check}

Our study proposes MREF-AD, which enables adaptive, interpretable fusion of Alzheimer's multimodal data by explicitly treating mesoscopic brain regions as distinct experts. 
Evaluated on ADNI benchmark data, the model demonstrates its strength not only in improved classification performance but also in handling missing modality and promoting clinical explainability into the disease process.
Indeed, one significant advantage of MREF-AD is its intrinsic interpretability.
Unlike black-box deep learning models or post-hoc analysis approaches~\cite{yi2023xgboost}, MREF-AD's gating weights provide a direct, probabilistic measure of feature importance.
The gating mechanism allows dynamic weights adjustment for different subjects, facilitating personalized interpretability.

Another key finding from our experiments is MREF-AD's stability under missing-MRI and -PET conditions, especially in contrast to Logistic Regression's significantly larger drop in performance. Interestingly, Logistic Regression has competitive performance under full-modality. This behavior has also been observed in previous works, where linear models often perform well in healthcare settings under complete data assumptions \cite{cabanillas2025evaluation-of-machine-learning-modles-for-the-prediction-of-alzheimers} compared with neural networks like MLP \cite{cary2021machine-learning-algorithms-to-predict-mortality}. Hence, the performance degradation under missing-modality scenarios highlights a key limitation that such models lack the architectural ability to handle missing data and rely on the quality of the imputation. While various modality imputation strategies have been explored, recent work increasingly emphasizes architecture-level approaches that inherently accommodate missing modalities \cite{wu2024deep-multimodal-learning-with-missing-modality-a-survey}. Thus, our results further validate the importance of considering architectural robustness to incomplete modalities for clinical deployment, where fully observed multimodal data are rarely guaranteed, and the need to move beyond complete data assumptions.
% \vspace{-3pt}

Beyond robustness to missing data, our results suggest that the modality-region expert decomposition of MREF-AD offers meaningful insights into AD neuroimaging. 
Although MREF-AD and the MLP baseline share similar neural network building blocks, MREF-AD consistently improves F1 and accuracy even in sparse settings, indicating that the gains are not simply driven by model size. 
Instead, the gating mechanism  captures heterogeneity in how different brain regions and modalities contribute to diagnosis across individuals.
The aggregated expert contributions from MRI and amyloid PET experts align with the clinical patterns of neurodegenerative atrophy and molecular pathology.
The modality ablations on how the model uses complementary information confirm the best performance arises from joint molecular and structural evidence.
The way MREF-AD approaches these patterns at a mesoscopic level also matches how clinicians often summarize AD-related abnormalities in practice.
% \vspace{-3pt}

Despite its promise, this study has several limitations. 
First, while we utilized ROI-level tabular data to ensure lightweight efficiency (e.g., as shown in Table~\ref{tab:complexity}), it discards voxel-level information and spatial patterns that raw image-based models could leverage.
Future work could replace the MLP experts in the MoE with 3D-CNN or visual transformer~\cite{liu2023survey} blocks to learn directly from the images.
Moreover, here we use only one imaging session per subject (last available PET/MRI visit), whereas future work could utilize longitudinal data for progression trajectories modeling or early detection, which is more important for clinical intervention of AD.
Finally, although assessed on MRI, Amyloid PET and demographic data as a proof-of-concept here, MREF-AD could be extended further to incorporate additional modalities such as tau PET, FDG PET, cognitive scores, CSF markers, and genetic data, to allow a more comprehensive view of AD etiology.

\section{Conclusions}
\label{sec:conclusions}

We presented MREF-AD, an adaptive Mixture-of-Experts framework for multimodal neuroimaging-based Alzheimer's disease diagnosis. By modeling amyloid PET, MRI, and demographic features as independent experts and using a gating network for subject-specific fusion, MREF-AD achieves robust and interpretable predictions. Furthermore, our results show that the model maintains AD diagnostic utility even when a specific imaging modality is missing at inference. Compared with traditional classifiers and state-of-the-art frameworks, MREF-AD achieved superior diagnostic performance across accuracy and F1-score, demonstrating the benefit of regional adaptive fusion. Beyond performance gains, the interpretability analysis revealed biologically meaningful expert weighting patterns, including modality--region combinations that align with known biomarker progression and highlight temporal and subcortical involvement in AD. These results highlight MREF-AD's potential as both a predictive and explanatory framework for multimodal neuroimaging. While we illustrate its utility on Alzheimer's disease diagnosis, the architecture is general and can be extended to other neurological disorders, additional imaging or molecular modalities, and longitudinal or prognostic modeling tasks.

\section*{Acknowledgments}

This work was supported in part by the NIH grants U01 AG066833, U01 AG068057, P30 AG073105, R01 AG068191, R01 AG07147, U19 AG074879, and R01 EB037101. 
% This work was supported in part by the following NIH grants: \textcolor{red}{TODO}. 
The ADNI data were obtained from the Alzheimer's Disease Neuroimaging Initiative (https://adni.loni.usc.edu), funded by NIH grant U01 AG024904. The authors declare no competing interests.

\vspace{-10pt}
\bibliographystyle{IEEEtran}
\bibliography{IEEEabrv, refs}

% \vspace{12pt}
\color{red}
% IEEE conference templates contain guidance text for composing and formatting conference papers. Please ensure that all template text is removed from your conference paper prior to submission to the conference. Failure to remove the template text from your paper may result in your paper not being published.

\end{document}